# Visual stream connectivity predicts assessments of image quality

Elijah F. W. Bowen[1*], Antonio M. Rodriguez[1], Damian Sowinski[1], Richard Granger[1]

[1] Dartmouth
6207 Moore Hall
Hanover, NH 03755

[*] Corresponding author
E-mail: efwb001@gmail.com

Declarations of interest: none

## ABSTRACT

Some biological mechanisms of early vision are comparatively well understood, but they have yet to be evaluated for their ability to accurately predict and explain human judgments of image similarity. From well-studied simple connectivity patterns in early vision, we derive a novel formalization of the psychophysics of similarity, showing the differential geometry that provides accurate and explanatory accounts of perceptual similarity judgments. These predictions then are further improved via simple regression on human behavioral reports, which in turn are used to construct more elaborate hypothesized neural connectivity patterns. Both approaches outperform standard successful measures of perceived image fidelity from the literature, as well as providing explanatory principles of similarity perception.

## 1 INTRODUCTION

How are subtle changes to an image registered in the brain? Images warped by various means such as lossy compression (e.g. JPEG, ISO-10918 [1] [2], explored herein) are recognizable, but appear measurably different to human observers. Perception of a degraded (e.g. compressed) image is not a linear function of the mean change to pixel luminance, nor any univariate function of mean change [3–12]. **Fig 1** illustrates a sample discrepancy between simple vector distances and perceived difference.

A set of distances forms a similarity space. It has been hypothesized that the formation of similarity spaces often found in neural codes [6] [13–15] may be a primary outcome of perception [16] [17]. In support this, perceptual similarity spaces have been valuably used to progress psychological theories of object and shape perception [16] [18–22] (to name a few).

Perceived similarity of degraded images has been the focus of the field of full-reference image quality assessment (IQA), whose approaches range from behavioral measures to image region weighting methods modeling saliency or attention [23–33] to some initial considerations of neural connectivity [4–10] [34], as well as just-noticeable differences (JNDs) and luminance masking [35–42]. Among the most-cited works is the Structural Similarity (SSIM) measure [12] [43], which combines metrics of pixel luminance, local contrast, and local correlation in normalized images.

Human judgments of similarity, which provide an approximation of the internal similarity space [13] [44], are a common and accessible psychophysical measurement. In principle, the anatomic structure and physiologic function of brain circuits must be carrying out these subjective visual operations. However, despite our unrivaled knowledge of the mammalian visual system, it can be very difficult to educe what underlying neural principles may be giving rise to a measured perceptual similarity.

Perceptual similarity need not be constrained to Euclidean metrics in image space [5–7] [9] [10] [23] [45]. To the contrary, several studies have concluded that aspects of visual perception are best described by non-Euclidean metrics [6] [7] [9] [34]. Returning to first principles, we provide an unexpectedly simple approach, in which straightforward properties of early visual pathway circuitry directly produce non-Euclidean similarity measures. The relations we find between stimulus and percept are in the tradition of psychophysics (e.g. [6] [46]) and IQA (e.g. [8]), but are distinguished by a high-dimensional strain model that naturally describes specific biological characteristics.

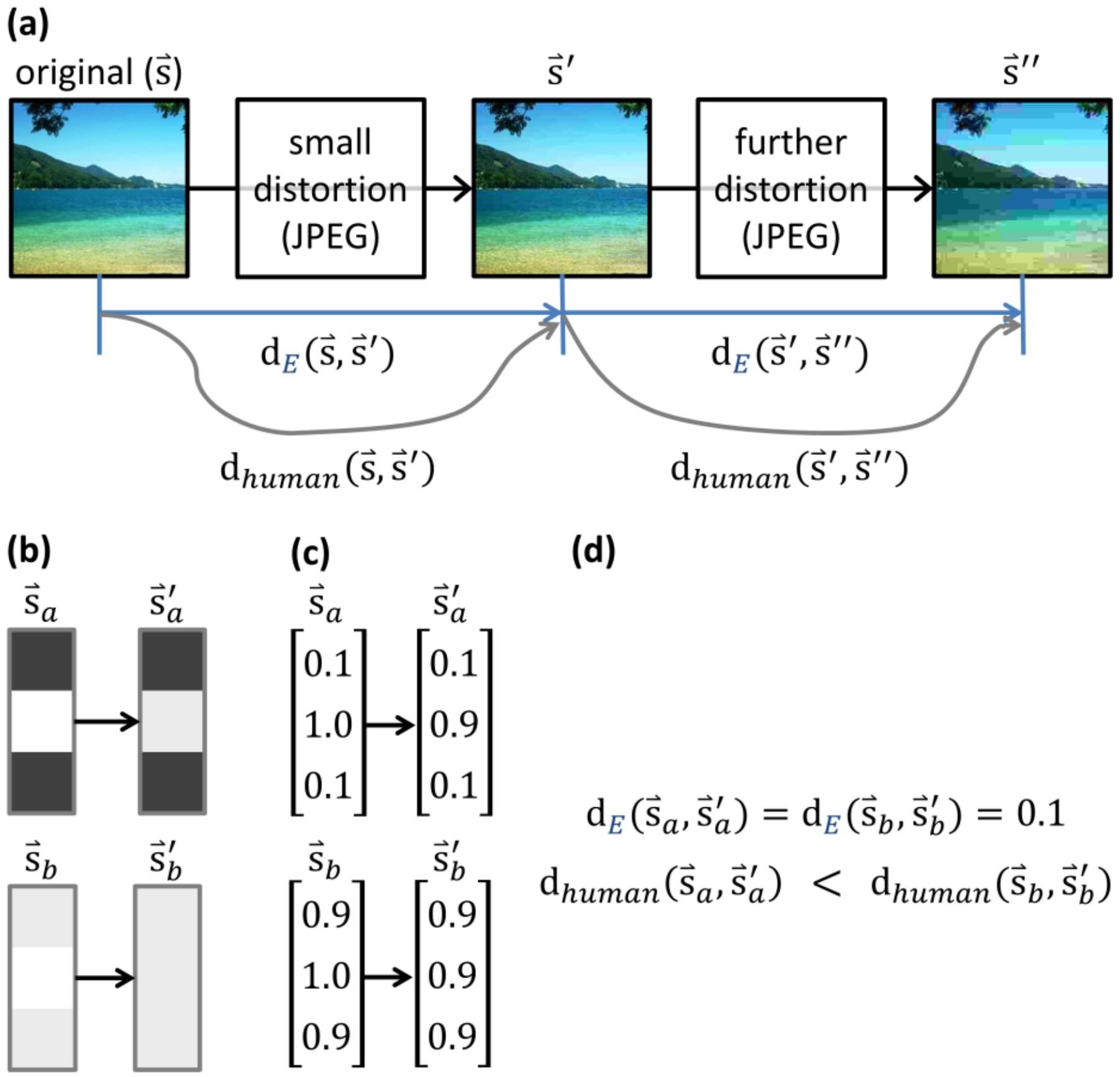

**Fig 1.** **(a)** As an original (non-degraded) image $\vec{s}$ becomes increasingly compressed (via a lossy method such as JPEG), how dissimilar are the images judged to be? Equal physical changes, in terms of average luminance, will not be perceived as equal to humans. **(b)** Enlarged three-pixel images $\vec{s}_a$ and $\vec{s}_b$, with degraded counterparts $\vec{s}'_a$ and $\vec{s}'_b$. **(c)** We can convert each image into a pixel vector of luminances (zero for black, one for white), as is often done. Euclidean distances can be computed between a pair of such image vectors by measuring luminance difference across each row, then combining the results. **(d)** The Euclidean distance between original and degraded images in both images is 0.1. However, humans overwhelmingly perceive change to be greater in the second ($\vec{s}_b \rightarrow \vec{s}'_b$) case, presumably due to the context of surrounding pixels.

## 2 THEORETICAL TREATMENT

### 2.1 A differential geometry of image perception

Bitmaps encode images on Cartesian coordinates of pixels – each axis denotes one (independent) pixel in an image, and the value along an axis is the luminance of that pixel. In the eye, photoreceptors possess a similar code. To calculate the dissimilarity between two image stimuli, $\vec{s}$ and $\vec{s}'$, a straw-man approach is to simply assume that the change in each bitmap pixel is independently processed, and then averaged (as in mean squared error):

$$\frac{1}{D}\sum_{d=1}^{D}(s'_d - s_d)(s'_d - s_d) \tag{1}$$

where $D$ is the number of pixels in the image. This can be rewritten, in linear algebra terms, as the dot product of the vector between $\vec{s}$ and $\vec{s}'$. Dropping the $1/D$ term (which is constant in each dataset) yields a measure of image difference which is the (squared) Euclidean distance:

$$\mathrm{d}_E^2(\vec{s},\vec{s}') = (\vec{s}' - \vec{s})^{\mathrm{T}}(\vec{s}' - \vec{s}) \tag{2}$$

In psychophysics, the relationship between Euclidean and perceived distance is often complicated, but consistent. This affords an opportunity to model perception as a structured deviation from Euclidean distance [6–10] [34]. We forward an analysis in which this neural transformation is modeled under continuum mechanics [47] as a displacement of images $\vec{s}$ to new "perceived" positions $\vec{s}_{\mathcal{P}}$ within image space. Each new position differs from the original via displacement field $\vec{u}(\vec{s})$:

$$\vec{s}_{\mathcal{P}} = \vec{s} + \vec{u}(\vec{s}) \tag{3}$$

Perception strains image space, which changes the Euclidean distances between stimuli (**Fig 2** lays out the problem in pixilated image space, where each dimension is the luminance of one pixel). The perceived difference between $\vec{s}$ and $\vec{s}'$ (the Euclidean distance between $\vec{s}_{\mathcal{P}}$ and $\vec{s}'_{\mathcal{P}}$) is:

$$\mathrm{d}_{\mathcal{P}}^2(\vec{s},\vec{s}') = (\vec{s}'_{\mathcal{P}} - \vec{s}_{\mathcal{P}})^{\mathrm{T}}\,(\vec{s}'_{\mathcal{P}} - \vec{s}_{\mathcal{P}}) \tag{4}$$

We seek to formalize the geometry of perceptual space – how points compare to one another. Geometry is agnostic to the exact value of this new position $\vec{s}_{\mathcal{P}}$. Therefore, we will focus on constructing a new measure of image difference which is a function of the original images (not of $\vec{s}_{\mathcal{P}}$). We now show that the displacement in eqn. (3) can be reinterpreted as distortions of the metric of the space in which the images live. Each difference vector between a perceived image and a perceived degraded copy is defined as:

$$\vec{s}'_{\mathcal{P}} - \vec{s}_{\mathcal{P}} = \vec{s}' + \vec{u}(\vec{s}') - \vec{s} - \vec{u}(\vec{s}) \quad (5)$$

Taylor expanding to first order, eqn. (5) reads as a function of how the displacement field has changed between $\vec{s}$ and $\vec{s}'$ (how strain changes as images change, $\vec{\nabla}_s\vec{u}$):

$$\vec{s}'_{\mathcal{P}} - \vec{s}_{\mathcal{P}} = (\vec{s}' - \vec{s}) + \left(\vec{\nabla}_s\vec{u}^{\mathrm{T}}\right)^{\mathrm{T}}(\vec{s}' - \vec{s}) \quad (6)$$

(This first-order approximation is only valid among points that are close enough that perceptual curvature is miniscule. While the degradations we evaluate (see **section 4**) are at times obvious to subjects, they are miniscule on the scale of image space - degradations never transform scenes into different scenes or make nonlocal changes. Secondly, the perceptual operators evaluated in this paper do not have extreme curvature. Finally, even an imprecise first-order approximation appears to capture valuable patterns.)

Eqn. (6) can be factored as:

$$\vec{s}'_{\mathcal{P}} - \vec{s}_{\mathcal{P}} = \left(\mathbf{I} + \left(\vec{\nabla}_s\vec{u}^{\mathrm{T}}\right)^{\mathrm{T}}\right)(\vec{s}' - \vec{s}) \quad (7)$$

where $\mathbf{I}$ is the identity matrix (the tensor of Cartesian coordinates in Euclidean space). $\vec{\nabla}_s\vec{u}^{\mathrm{T}}$ is a matrix where the value in the $i$th row and $j$th column, $\partial u_i/\partial s_j$, describes how much additional displacement the luminance change to pixel $j$ contributes to the perceptual displacement of pixel $i$:

$$\left(\vec{\nabla}_s\vec{u}^{\mathrm{T}}\right)^{\mathrm{T}} \equiv \begin{bmatrix} \frac{\partial u_1}{\partial s_1} & \cdots & \frac{\partial u_1}{\partial s_D} \\ \vdots & \ddots & \vdots \\ \frac{\partial u_D}{\partial s_1} & \cdots & \frac{\partial u_D}{\partial s_D} \end{bmatrix} \quad (8)$$

The Euclidean distance metric in Cartesian coordinates of pixels (e.g. eqns. (2) and (4)) has the identity matrix as its tensor. Now that we understand eqn. (7), we can compute perceived distance (eqn. (4)) without reference to $\vec{s}_{\mathcal{P}}$:

$$\mathrm{d}^2_{\mathcal{P}}(\vec{s}, \vec{s}') = (\vec{s}' - \vec{s})^{\mathrm{T}}\left(\mathbf{I} + \vec{\nabla}_s\vec{u}^{\mathrm{T}}\right)\mathbf{I}\left(\mathbf{I} + \left(\vec{\nabla}_s\vec{u}^{\mathrm{T}}\right)^{\mathrm{T}}\right)(\vec{s}' - \vec{s}) \quad (9)$$

See **section 3.1** for a derivation. It will be useful to define $\mathbf{J}$, the Jacobian of the perceptual distortion:

$$\mathbf{J} = \mathbf{I} + \left(\vec{\nabla}_s \vec{\mathrm{u}}^{\mathrm{T}}\right)^{\mathrm{T}} \tag{10}$$

Therefore, the perceived difference between $\vec{\mathrm{s}}$ and $\vec{\mathrm{s}}'$ is simply:

$$\mathrm{d}_{\mathcal{P}}^2(\vec{\mathrm{s}}, \vec{\mathrm{s}}') = (\vec{\mathrm{s}}' - \vec{\mathrm{s}})^{\mathrm{T}} \mathbf{J}^{\mathrm{T}} \mathbf{I}\, \mathbf{J}\, (\vec{\mathrm{s}}' - \vec{\mathrm{s}}) \tag{11}$$

where image difference has been distorted by perceptual strain. If there exists no perceptual strain, $\mathbf{J} = \mathbf{I}$, and $\mathrm{d}_{\mathcal{P}}^2(\vec{\mathrm{s}}, \vec{\mathrm{s}}') = \mathrm{d}_E^2(\vec{\mathrm{s}}, \vec{\mathrm{s}}')$.

## 2.2 Mathematical relationship between biological projection patterns and perceptual strain

We define perception $\mathbf{P}$ as an operation which changes each bitmap image stimulus $\vec{s}$ (a point on Cartesian coordinates of pixels) to a perceived vector $\vec{s}_{\mathcal{P}}$:

$$\vec{s}_{\mathcal{P}} = \mathbf{P}\,\vec{s} \tag{12}$$

Let us say that $\mathbf{P}$ is a locally multilinear operator - a $D$ x $D$ matrix for each stimulus. The main diagonal represents 1:1 topographic connectivity among neurons, or an unmodified percept. Each off-diagonal describes an additional biological connection or perceptual interaction (that may strain image space). This quantification of perception is very natural to psychophysicists and neuroscientists. Moreover, it is a simple way to quantify connectomes and local projection patterns like those in **Fig 3** - each element of $\mathbf{P}$ is a scalar function describing how a pair of neurons, receptive fields, concepts, or brain regions relate.

In the retina, image pixels are topographically mapped to photoreceptors - adjacent pixels are processed by adjacent photoreceptors. In turn, neighboring photoreceptors project to neighboring retinal cells, with some lateral excitation and inhibition (**Fig 3a**), so that a given retinal cell receives information from a small contiguous region of an image [48]. The resulting receptive fields are typically fit by a Gaussian function of relative retinotopic position between points $i$ and $j$ on an image (retinal distance, $\mathrm{d}_{ret}(i,j)$) [48–52] (for electrophysiological examples, see **Fig 3b**):

$$\mathrm{Gauss}(\mathrm{d}_{ret}(i,j)) = \exp\left(\frac{-\mathrm{d}_{ret}(i,j)^2}{2\sigma^2}\right) \tag{13}$$

The early visual stream conserves this topographic connectivity [48] [53–61] yielding Gaussian and center-surround connectivity. The latter can be approximated as the difference between two concentric Gaussian functions of distance – a narrow center and broad surround (difference of Gaussians, DOG; **Fig 3c**):

$$\mathrm{DOG}(\mathrm{d}_{ret}(i,j)) = \frac{1}{1+\alpha}\exp\left(\frac{-\mathrm{d}_{ret}(i,j)^2}{2\sigma_{center}^2}\right) - \frac{\alpha}{1+\alpha}\exp\left(\frac{-\mathrm{d}_{ret}(i,j)^2}{2\sigma_{surround}^2}\right) \tag{14}$$

Eqns. (13) and (14) describe how pixels will be combined when these cells process an image. Somewhat independently, psychophysicists often measure scalar differences between end percepts. By connecting neuroscience to psychophysics, we can begin to generate new predictions and understandings of how underlying biology explains behavior. We are ready to specify how perceptual strain links these two bodies of knowledge. We set the biological projection in eqn. (12) equal to continuum mechanics eqn. (3) (and rearrange), yielding:

$$\vec{u}(\vec{s}) = \mathbf{P}\vec{s} - \vec{s} \tag{15}$$

However, to measure distance, we wish to relate $\mathbf{P}$ to the *gradient* of $\vec{u}$:

$$\vec{\nabla}_s \vec{u}^{\mathrm{T}} = \vec{\nabla}_s\left(\vec{s}^{\mathrm{T}} \mathbf{P}^{\mathrm{T}} - \vec{s}^{\mathrm{T}}\right) \tag{16}$$

The gradient of $\vec{s}$ with respect to itself is simply $\mathbf{I}$. We make this replacement and apply a transpose, returning the left side to something more recognizable:

$$\left(\vec{\nabla}_s \vec{u}^{\mathrm{T}}\right)^{\mathrm{T}} = \mathbf{P} - \mathbf{I} \tag{17}$$

Eqn. (17) shows that any known perceptual operator $\mathbf{P}$ can be written in terms of the derivative of the displacement field - we can say that the displacement field is *generated by* the perceptual operator.

## 2.3 Predicting perceived distance

We evaluate several possible forms of $\mathbf{P}$ herein. The first is Gaussian connectivity between the cells that favor pixels $i$ and $j$ (as described earlier and in **Fig 3b**):

$$p_{i,j_{Gauss}} = \mathrm{Gauss}\big(\mathrm{d}_{ret}(i,j)\big) \tag{18}$$

For a perceptual operator using difference of Gaussians, a simple change to eqn. (18) suffices:

$$p_{i,j_{DOG}} = \mathrm{DOG}\big(\mathrm{d}_{ret}(i,j)\big) \tag{19}$$

For convenience, we designed the Gaussian form in eqn. (13) to provide $\mathrm{Gauss}(0) = 1$. In fact, when choosing our units, we set all perceptual relations relative to the center of the receptive field. Each $\mathbf{P}$ has $1$'s along the diagonal. When we subtract $\mathbf{I}$ in eqn. (17), we zero the diagonal elements of $(\vec{\nabla}_s\vec{\mathrm{u}}^{\mathrm{T}})^{\mathrm{T}}$ (and thus the strain tensor; see **section 3.2**). Together, these components account for image space dilation, which cannot be measured using relative psychophysical distances. (Diagonal connectivity - the Euclidean component of perception - is separately represented by $\mathbf{I}$ in eqn. (10).)

It will be seen in **section 5** that both Gaussian and DOG versions of this equation, with no further modifications, provide unexpectedly accurate predictions of human image similarity judgments, in fact very surprisingly outperforming approaches that are designed specifically for the task (**Table 1**, **Fig 7**).

Eqn. (17) is a simple way to generate the perceptual displacement field from $\mathbf{P}_{Gauss}$. or $\mathbf{P}_{DOG}$. This simple approach ("approach I") produces a Jacobian from the displacement field, which lets us measure the perceived distance between two stimuli. The resulting Jacobians are of course unlikely to be perfectly accurate representations of the actual connectivity patterns in early visual pathways, which are shaped by development and learning. Thus, in a second approach ("approach II"), we regress on pairs of image change $(\vec{\mathrm{s}}' - \vec{\mathrm{s}})$ and human evaluations of dissimilarity. We vary each cell of the perceptual Jacobian until the resulting tensor produces image dissimilarities that are locally maximally Pearson-correlated with human ratings. The resulting Jacobian may be hypothesized to more accurately correspond to the transforms that may be taking place along the connections in the early visual pathway, as in **Fig 3d**. Alternately, this approach may be considered roughly accordant with methods in the IQA literature that attempt to learn optimal predictors of human judgments (e.g. [62] [63]).

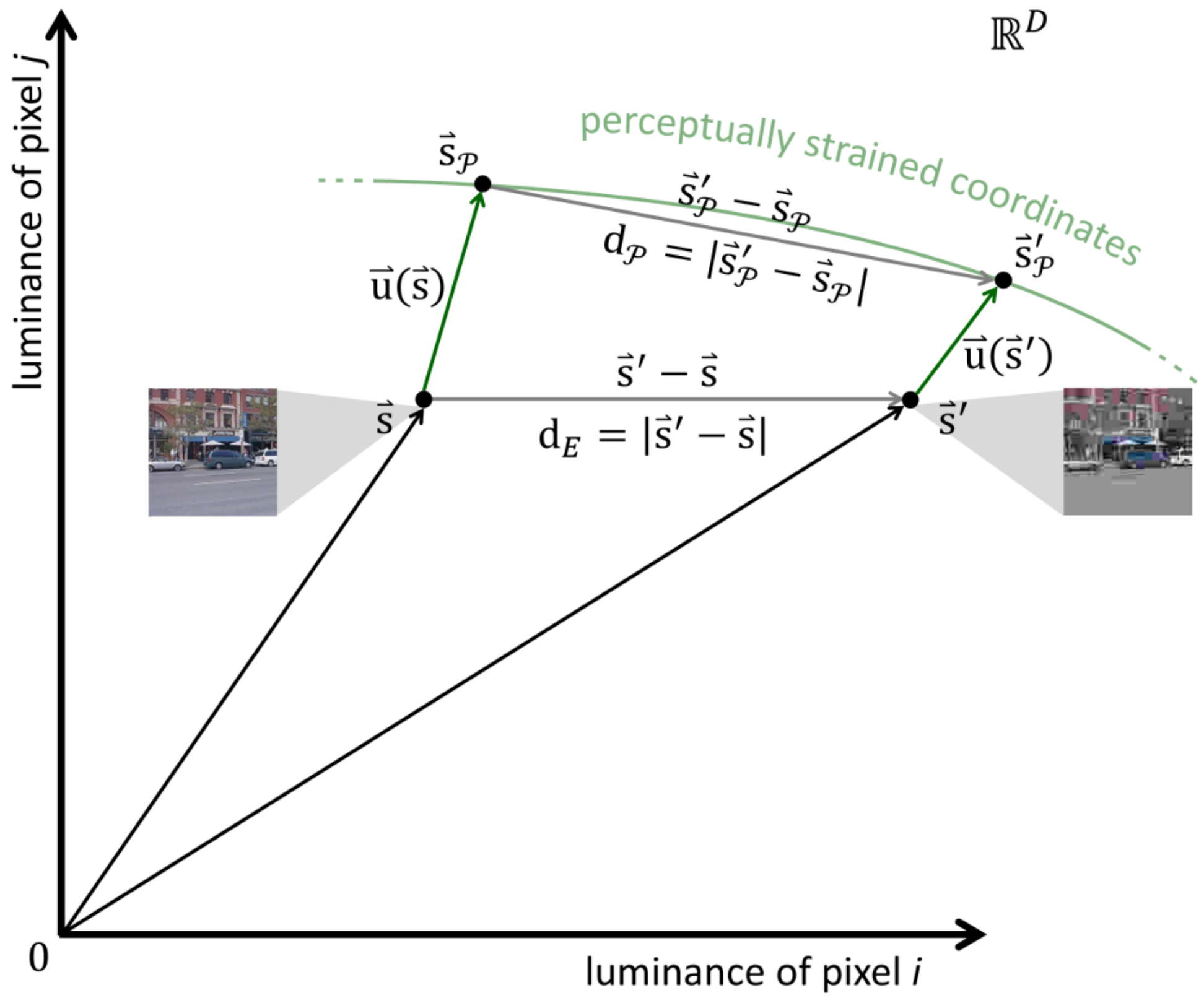

**Fig 2. Vector space account of perceptual strain**. Each possible image can be considered a point (i.e. a vector from the origin, black arrows) in pixelated image space, where each Cartesian coordinate is the luminance of one pixel. Here we plot only two such coordinate axes for simplicity. When humans perceive images, cells form population codes which carefully change the representations of the light patterns. Therefore, an image $\vec{s}$ and its degraded counterpart $\vec{s}'$ are displaced to new coordinates $\vec{s}_{\mathcal{P}}$ and $\vec{s}'_{\mathcal{P}}$. This perceptual strain is quantified as a vector field $\vec{u}(\vec{s})$ which can be evaluated at any image (green arrows). Approach I defines $\vec{u}(\vec{s})$ in terms of biological connectivity patterns. Approach II triangulates the vector field of perceptual strain from Euclidean ($d_E$) and perceived ($d_{\mathcal{P}}$) distance measurements. Crucially, in our hands perceived distance is Euclidean after the correct perceptual displacement field is applied to images. The new image positions $\vec{s}_{\mathcal{P}}$ are left as an internal property of neural representations, not relevant to approximate on these coordinates.

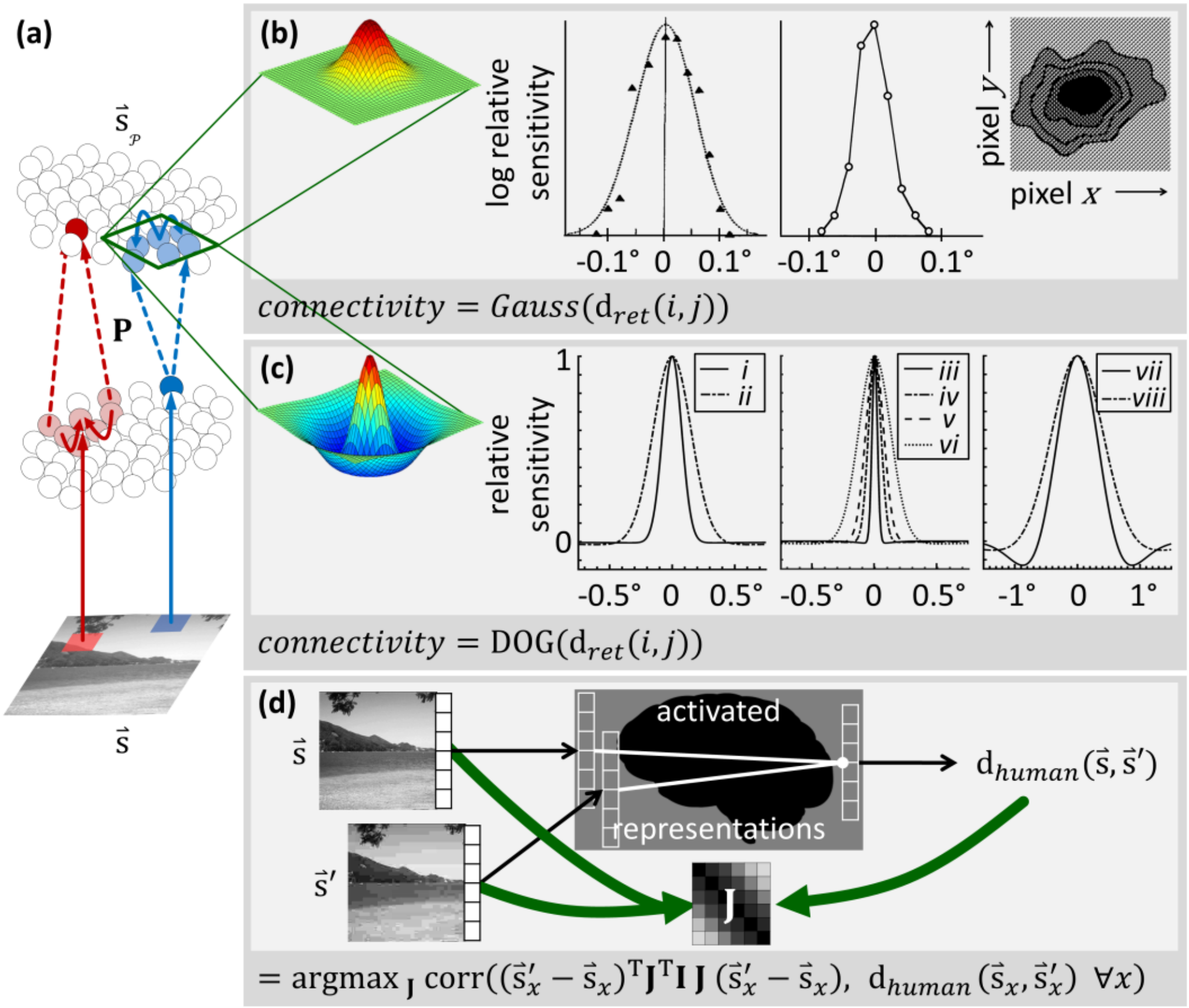

**Fig 3. Principles formalized from neural connectivity. (a)** Caricature of neural projections from cells in the lower population (where each cell represents light at one location, much like pixels), to downstream cells, with a topographic connectivity pattern. Neighboring cells represent neighboring pixels. These cells connect with their neighbors (solid arrows). Downstream, cells again receive projections from neighbors (dashed arrows). See text. **(b)** Approach I, with a Gaussian connectivity function (see text). Such connectivity is often found in (left to right) retinal ganglion cells [51] [50] and visual cortex [52]. **(c)** Approach I, with a difference-of-Gaussian connectivity function (see text). Such connectivity is found in, for example, OFF-center retinal bipolar cells (***i*** midget, ***ii*** diffuse [48]). Similar connectivity is also found downstream in retinal ganglion cells which project to the parvocellular (***iii*** 0°-5°, ***iv*** 10°-20° from visual center [53]) or magnocellular (***v*** 0°-10°, ***vi*** 10°-20° from visual center [53]) pathways, across the visual field (***vii*** peripheral [54] [55], ***viii*** <10° from fixation [60]). **(d)** Using approach II, we can model the pattern of connectivity which, given simplicity assumptions, best explains how images relate to human ratings. Our objective is a Jacobian **J** which causes maximal correlation between perceptual dissimilarity (computed between each $\vec{s}$ and $\vec{s}'$ using eqn. (11)) and human difference ratings (see text).

## 3 CALCULATION

### 3.1 Derivation of perceptual distance

In this paper, we defined $\vec{u}(\vec{s})$ as the distortion field which places images $\vec{s}$ in new locations $\vec{s}_{\mathcal{P}}$ where, locally, perceived difference matches Euclidean distance. We wrote that eqn. (2), $d_E^2(\vec{s}, \vec{s}') = (\vec{s}' - \vec{s})^T(\vec{s}' - \vec{s})$, can be meaningfully converted into eqn. (9), $d_{\mathcal{P}}^2(\vec{s}, \vec{s}') = (\vec{s}' - \vec{s})^T(\mathbf{I} + \vec{\nabla}_s\vec{u}^T)\mathbf{I}(\mathbf{I} + (\vec{\nabla}_s\vec{u}^T)^T)(\vec{s}' - \vec{s})$. Let us derive this using **Fig 2**. The difference between perceptual image coordinates is given by eqn. (4), $d_{\mathcal{P}}^2(\vec{s}, \vec{s}') = (\vec{s}'_{\mathcal{P}} - \vec{s}_{\mathcal{P}})^T\, (\vec{s}'_{\mathcal{P}} - \vec{s}_{\mathcal{P}})$. By applying eqn. (3) (or by using a little trigonometry on **Fig 2**):

$$d_{\mathcal{P}}^2(\vec{s}, \vec{s}') = \left(\vec{s}' + \vec{u}(\vec{s}') - \vec{s} - \vec{u}(\vec{s})\right)^T \left(\vec{s}' + \vec{u}(\vec{s}') - \vec{s} - \vec{u}(\vec{s})\right) \tag{20}$$

Importantly, can replace $\vec{u}(\vec{s}')$ with $\vec{u}(\vec{s})$ *plus* the degree to which $\vec{u}(\vec{s}')$ differs from $\vec{u}(\vec{s})$ as we move from $\vec{s}$ to $\vec{s}'$:

$$\vec{u}(\vec{s}') = \vec{u}(\vec{s}) + \left(\vec{\nabla}_s\vec{u}^T\right)^T(\vec{s}' - \vec{s}) \tag{21}$$

We apply this replacement to eqn. (20), then cancel the $\vec{u}(\vec{s}) - \vec{u}(\vec{s})$ terms. This yields a function only of images and changes to $\vec{u}$ as a function of changes to pixel intensity:

$$d_{\mathcal{P}}^2(\vec{s}, \vec{s}') = \left(\vec{s}' + \left(\vec{\nabla}_s\vec{u}^T\right)^T(\vec{s}' - \vec{s}) - \vec{s}\right)^T \left(\vec{s}' + \left(\vec{\nabla}_s\vec{u}^T\right)^T(\vec{s}' - \vec{s}) - \vec{s}\right) \tag{22}$$

We reorder this equation, then distribute the outer transpose:

$$d_{\mathcal{P}}^2(\vec{s}, \vec{s}') = \left((\vec{s}' - \vec{s})^T + (\vec{s}' - \vec{s})^T\, \vec{\nabla}_s\vec{u}^T\right)\left((\vec{s}' - \vec{s}) + \left(\vec{\nabla}_s\vec{u}^T\right)^T(\vec{s}' - \vec{s})\right) \tag{23}$$

Next, we factor each half of the formula, returning the equation to a form we recognize - eqn. (9): $d_{\mathcal{P}}^2(\vec{s}, \vec{s}') = (\vec{s}' - \vec{s})^T(\mathbf{I} + \vec{\nabla}_s\vec{u}^T)\mathbf{I}(\mathbf{I} + (\vec{\nabla}_s\vec{u}^T)^T)(\vec{s}' - \vec{s})$.

We started with a simple difference measure in terms of unknown perceptual/neural representations, $(\vec{s}'_{\mathcal{P}} - \vec{s}_{\mathcal{P}})^T\, (\vec{s}'_{\mathcal{P}} - \vec{s}_{\mathcal{P}})$. This equation has now been converted into a strained difference measure in terms of image pixels.

### 3.2 Derivation of strain tensor

In the previous subsection, we could have replaced eqn. (9) with the inner multiplication:

$$\mathrm{d}_{\mathcal{P}}^2(\vec{\mathrm{s}},\vec{\mathrm{s}}') = (\vec{\mathrm{s}}' - \vec{\mathrm{s}})^{\mathrm{T}} \left(\mathbf{I} + \vec{\nabla}_s\vec{\mathrm{u}}^{\mathrm{T}} + \left(\vec{\nabla}_s\vec{\mathrm{u}}^{\mathrm{T}}\right)^{\mathrm{T}} + \vec{\nabla}_s\vec{\mathrm{u}}^{\mathrm{T}}\left(\vec{\nabla}_s\vec{\mathrm{u}}^{\mathrm{T}}\right)^{\mathrm{T}}\right)(\vec{\mathrm{s}}' - \vec{\mathrm{s}}) \tag{24}$$

The last term in the middle is an order smaller than the other terms. If we assume that it is vanishingly small, we reach a new equation:

$$\mathrm{d}_{\mathcal{P}}^2(\vec{\mathrm{s}},\vec{\mathrm{s}}') = (\vec{\mathrm{s}}' - \vec{\mathrm{s}})^{\mathrm{T}} \left(\mathbf{I} + \vec{\nabla}_s\vec{\mathrm{u}}^{\mathrm{T}} + \left(\vec{\nabla}_s\vec{\mathrm{u}}^{\mathrm{T}}\right)^{\mathrm{T}}\right)(\vec{\mathrm{s}}' - \vec{\mathrm{s}}) \tag{25}$$

We define the strain tensor based on the above equation:

$$\mathrm{d}_{\mathcal{P}}^2(\vec{\mathrm{s}},\vec{\mathrm{s}}') = (\vec{\mathrm{s}}' - \vec{\mathrm{s}})^{\mathrm{T}}(\mathbf{I} + 2\boldsymbol{\varepsilon})(\vec{\mathrm{s}}' - \vec{\mathrm{s}}) \tag{26}$$

where $\boldsymbol{\varepsilon}$ is the strain tensor and $\mathbf{I}$, the identity matrix, was the original tensor. By distribution of the previous equation, we can create an alternative definition of perceptual distance (not required herein):

$$\mathrm{d}_{\mathcal{P}}^2(\vec{\mathrm{s}},\vec{\mathrm{s}}') \rightarrow \mathrm{d}_{E}^2(\vec{\mathrm{s}},\vec{\mathrm{s}}') + 2\,(d\vec{\mathrm{s}})^{\mathrm{T}}\,\boldsymbol{\varepsilon}\,d\vec{\mathrm{s}} \tag{27}$$

where $2\,(d\vec{\mathrm{s}})^{\mathrm{T}}\,\boldsymbol{\varepsilon}\,d\vec{\mathrm{s}}$ is the *change in distance* caused by perceptual strain.

# 4 MATERIALS AND METHODS

## 4.1 Published datasets

We evaluate our perceptual model on the JPEG portion of three industry-standard datasets. The Categorical Subjective Image Quality (CSIQ) dataset [38] contains 30 hand-selected color 512 pixel x 512 pixel images of animals, landscapes, people, plants, and urban scenes. The images were degraded to five different levels of JPEG fidelity, which human subjects (N<35, precise count unknown) placed together on a linear scale such that pairwise distances between the images matched perceived difference. The TID 2013 [64] and Toyama [65] datasets were also utilized for breadth. These datasets contain similar imagery, with slightly varying image sizes and measures. See citations. Regardless of dataset, all human ratings reported here are normalized to a range of [0,1] (0 being no perceived distance; perfect fidelity).

Preexisting datasets have been shown to be inconsistent benchmarks, even when datasets contain almost the exact same images - CSIQ, TID 2013, and Toyama share many images, but prefer different IQA measures [66] [67]. Our work also raises the possibility of comparing natural image statistics with perceptual geometry, but such comparisons would require more plentiful and controlled imagery. We therefore introduce the new Scene Image Quality ("SceneIQ") dataset. Whereas CSIQ, TID 2013, and Toyama each contain less than 50 original (non-degraded) images, SceneIQ contains 2080 original images.

## 4.2 Newly acquired SceneIQ dataset

We acquired human fidelity ratings for a public set of 256 x 256 pixel color images [68], split into eight naturalistic scene categories: seacoast, forest, highway, inside city, mountain, open country, street, and tall building (for examples, see **Fig 4A**). The images were randomly subsetted from the original publication to equalize N across categories. We used 260 images per category, the number of images in the rarest category. Each image in the dataset was degraded into four JPEG quality levels: 30%, 20%, 10%, and 5% using ImageJ [69].

The dataset contains 260*8 = 2080 non-degraded images, and 8320 degraded images. This high count is important to reduce regression over-fitting and evaluate higher-order statistics, but poses a problem, because no single subject can rate this many images. While other assignment strategies were evaluated, we decided to split the subjects into groups. Each subject was randomly assigned to a set of 40 images, without replacement, such that every image is seen exactly once within one group of 52 people. We collected enough data for 5 groups, or 260 subjects from (primarily) American humans on Mechanical Turk. Each image was seen by 5 people, yielding a total of 41,600 ratings. While human ratings were based on color JPEGs, all IQA algorithms used grayscaled versions. This is the standard procedure in the field of IQA - measures such as SSIM cannot be computed on multispectral data. For all computer IQA measures, images were normalized (luminance stretched) to a range of 0-255.

In each randomly shuffled trial, one non-degraded image was presented per screen, along with the four degraded versions of the same image in random order. To make a series of pairwise comparisons, subjects could left click to magnify any thumbnail in the left box, and right click to magnify it in the right box (screenshot in **Fig 4B**). Subjects were instructed to rate each degraded image based on how different it was from the original, on an integer scale from 0 to 100 (instructions are available in **Supplemental Fig S1**). These ratings were converted to difference mean opinion scores (DMOS, a standardized measure) by the equation $\mathrm{DMOS} = 1 - (\mathrm{rating} \ / \ 100)$. A DMOS of 1 rates two images as 100% different (an undefined concept). A DMOS of 0 indicates that images have no perceptible difference. When subjects were satisfied with the correctness of all four ratings (without time constraint), they clicked "Accept" to advance to the next trial.

Ratings obtained via Mechanical Turk involve everyday viewing conditions, and thus are less controlled for viewing parameters. So, to validate/baseline these "SceneIQ Online" DMOS ratings obtained via Mechanical Turk, we compared with the commonly accepted DMOS scores of CSIQ. As indicated by **Fig 4C**, the original DMOS scores for the CSIQ dataset, collected in a controlled environment, have on average half the variance of those collected using our online paradigm. However, Mechanical Turk enables mass data collection, so we can use more subjects and more images to enhance power. This approach has been shown to improve statistical significance [70] (indeed, **Fig 4D** indicates small standard error bars in one simple analysis). Perhaps, the ecological validity of real, variable viewing conditions makes these ratings an even more reliable benchmark than scores collected under highly controlled conditions.

Five subjects were discarded and replaced with new ones:  Two were discarded because their data became corrupted during collection. Two were discarded because they self-reported as having poor vision. One was discarded for disregarding task instructions, almost always responding with the maximum rating value. To avoid biasing the dataset, we chose liberal inclusion criteria. To enable more strict exclusion of mechanical turks which poorly adhered to the task, future iterations of this dataset would benefit from catch trials or measures of task performance orthogonal to the dependent variables.

The paradigm and stimuli were replicated in a controlled laboratory setting at Dartmouth ("SceneIQ Lab" dataset). All subjects used the same high quality screen (Samsung U28D590), same private viewing room, and similar viewing distance (50.8 cm, 20 inches). 49 subjects (aged 18 – 22 years, 35 females) participated in rating degraded versions of 600 original (non-degraded) images (75 per semantic category), extracted from the same image source as SceneIQ [68] and degraded identically. 4 participants were discarded for incomplete data.

To eliminate uncontrolled inconsistencies between the CSIQ and SceneIQ datasets, we will also report a dataset ("CSIQ Revised") collected using CSIQ original (non-degraded) images but SceneIQ image degradation and methodology. Consistently with SceneIQ, we degraded each CSIQ image to four JPEG quality levels: 30%, 20%, 10%, and 5% using ImageJ [69]. 40 subjects each rated all images via Mechanical Turk using the same paradigm as SceneIQ Online. No subjects were discarded.

All human studies were approved by the Dartmouth institutional review board. Additional summary statistics are visible in **Supplemental Fig S2**. SceneIQ can be acquired online at https://github.com/DartmouthGrangerLab/SceneIQ or by contacting the authors. Additional methodological details can also be found online.

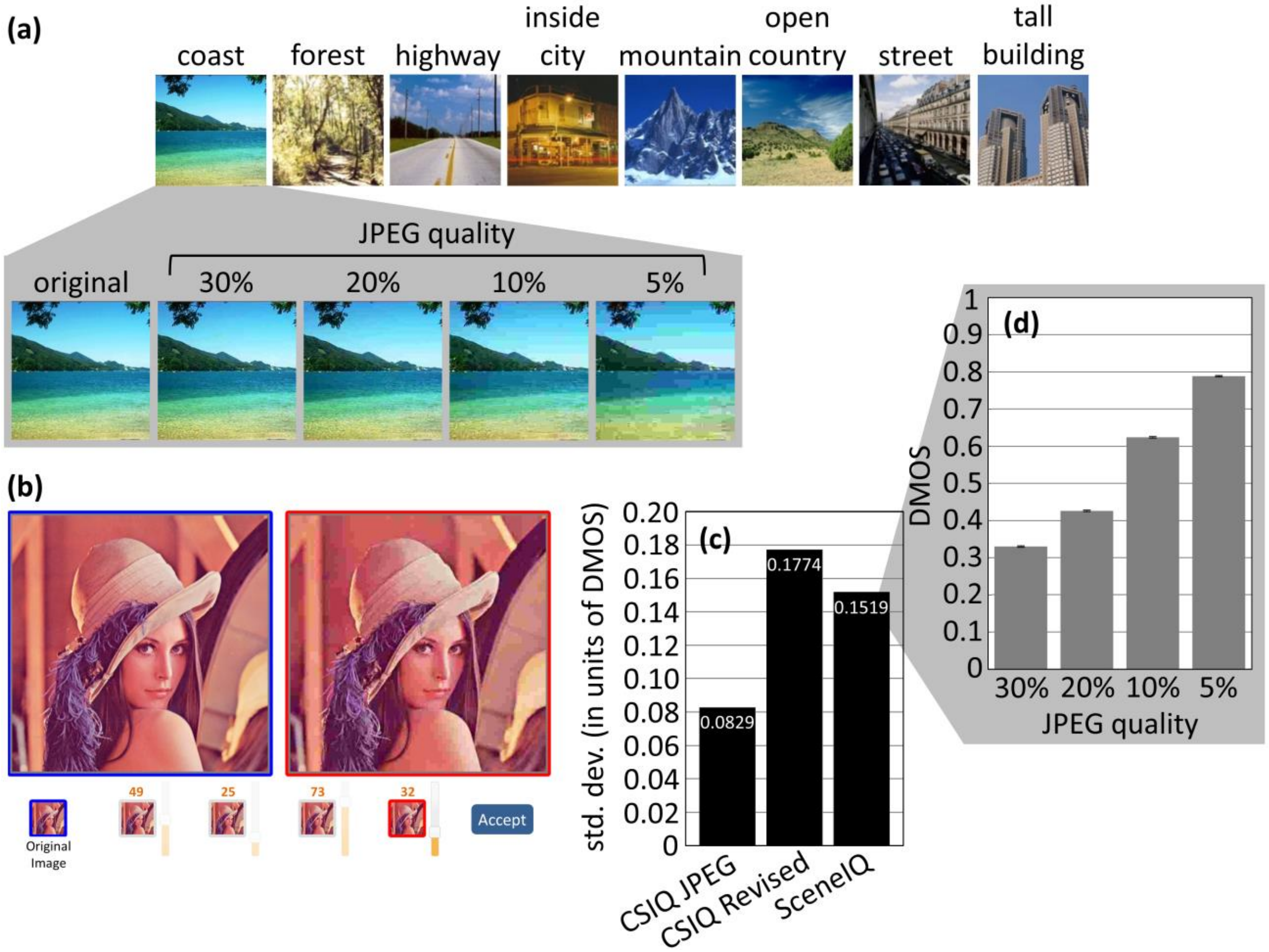

**Fig 4. Characteristics of the SceneIQ dataset. (a)** Example images. **(b)** Layout of the experimental paradigm. **(c)** Mean of the subject-wise standard deviation across all images and quality levels for three sets of data (left to right): the CSIQ (JPEG) dataset's original DMOS scores, DMOS scores for CSIQ non-degraded images degraded at the same quality levels used on the SceneIQ dataset and scored on Mechanical Turk, and the SceneIQ scores. **(d)** SceneIQ Online dataset. Mean DMOS score increases as the JPEG quality decreases (humans rate lower fidelity images as being lower fidelity). Bars are standard error across images (N=2080).

## 4.3 Approach I

The human perceptual operator was first described by a Gaussian function with a standard deviation of σ. With the exception of SceneIQ Lab, all of our datasets lack the viewing parameters needed to convert our stimulus model in terms of pixels into degrees visual angle. So, without loss of generality we define the retinal distance between two adjacent pixels as the unit distance. We also cannot define a σ from biological projection patterns without this conversion. Instead, given the Gaussian hypothesis, we seek the optimal parameterization of σ, and judge the hypothesis on its best terms. (Prior works have similarly optimized hypothesis parameters using real data [43] [71].) The model was evaluated with σ in the range [0.4,3.0] by increments of 0.1 pixels. Across five random folds of the original (non-degraded) images, we recorded the σ of the best correlations with DMOS (see **Fig 5** for error curves). We present further analyses using two sets of optimal parameters, σ = 2.0 pixels for CSIQ (JPEG) and σ = 0.9 pixels for SceneIQ. We did not note at the time that there may be some similarity between these values and the correlation among pixels recorded in the image dataset itself (see **Supplemental Fig S 9 B**).

The perceptual operation was next described by a difference between two concentric Gaussian functions. We evaluated a range of values for center Gaussian width (by increments of 0.2 pixels on the range [0.6,5.0] pixels), surround (negative) Gaussian width (increments of 0.2, range [0.6,5.6]; larger values become slow to compute), and the ratio of maximum heights between the two Gaussians, α (increments of 0.1, range [0.5,1.5]). Using the same crossvalidation procedure as before, we recorded Pearson correlation with SceneIQ Online DMOS for each possible parameter combination. All folds reported the same local maximum ($\sigma_{center}$ = 3.6 pixels, $\sigma_{surround}$ = 5.2 pixels, α = 0.7). The result for one fold is reported in **Fig 5G**.

Both connectivity patterns described in this section were derived from biology. However, their parameterizations were derived by model fit. This yields a high-performing hybrid Jacobian, but does not fully evaluate the relative contributions of actual Gaussian and difference-of-Gaussian connectivity profiles in the human. We look forward to further evaluations of approach I with stimulus-independent, biologically derived Gaussian widths.

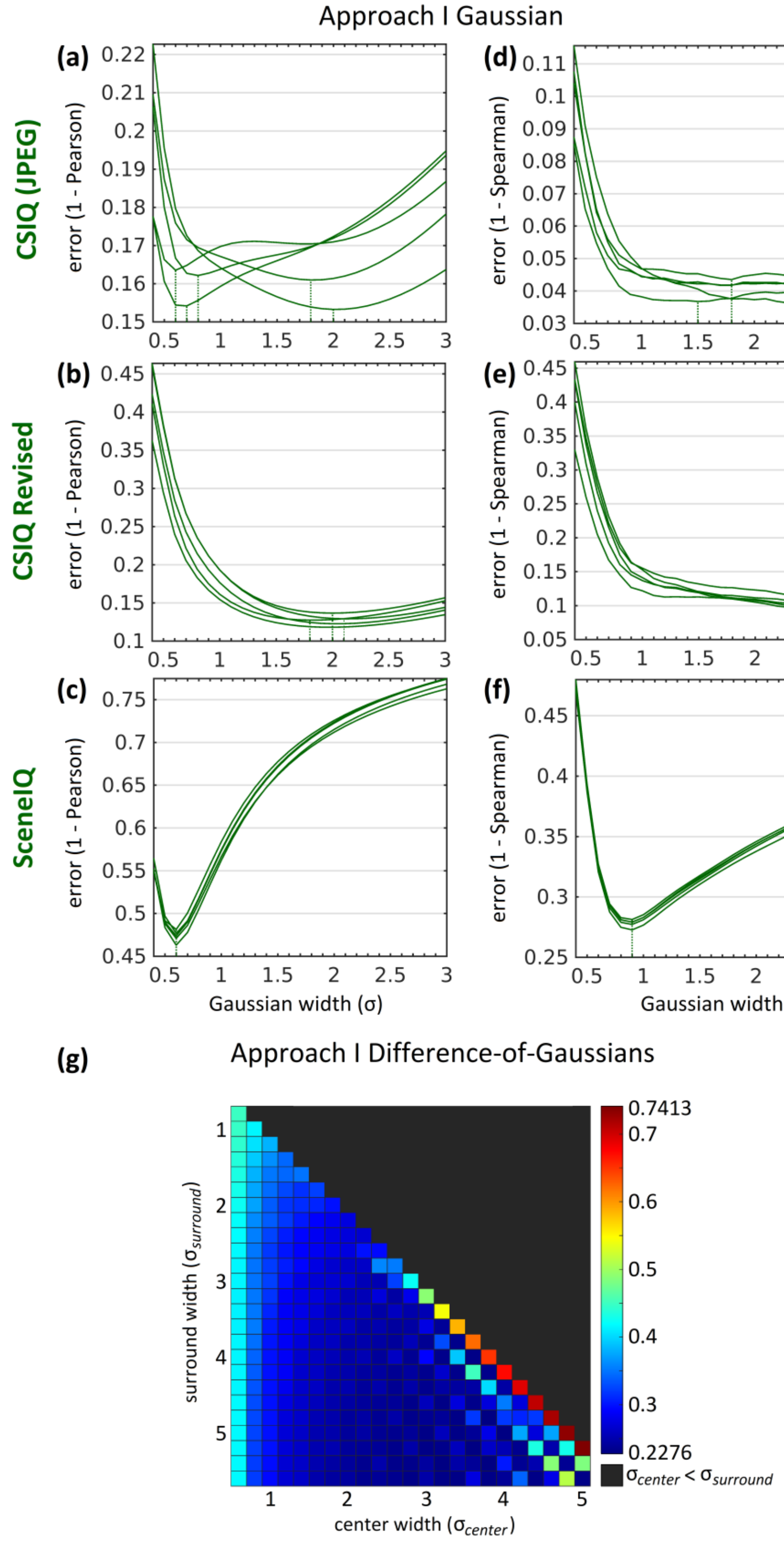

**Fig 5. Approach I optimality with various Gaussian widths.** A range of Gaussian widths ($\sigma$) were evaluated for each of five random folds of the **(a)** CSIQ (JPEG) dataset, **(b)** CSIQ Revised dataset, **(c)** SceneIQ Online dataset using Pearson correlation. **(d)** CSIQ (JPEG), **(e)** CSIQ Revised, and **(f)** SceneIQ Online were also evaluated using Spearman correlation. Dashed lines mark the global minima of each fold. **(g)** Difference-of-Gaussian training error as a combined function of $\sigma_{center}$ and $\sigma_{surround}$, on SceneIQ Online fold 1 of 2. For visualization, the third parameter ($\alpha$) was eliminated by selecting its optimal value for each combination of $\sigma_{center}$ and $\sigma_{surround}$.

## 4.4 Approach II

Our second approach uses regression to find a perceptual Jacobian that measures the distance between image pairs ($\vec{s}' - \vec{s}$) in a humanlike way - in correlation with DMOS scores. Treating our regression as an optimization problem, we performed random walk gradient descent on the set of all possible combinations of values for the cells of the Jacobian. The Jacobian was initialized to the identity matrix (the initial distance measure was Euclidean). At each iteration, the algorithm randomly selected a cell of the Jacobian. This corresponded to a particular dimension of the error surface. The algorithm then evaluated the error of the Jacobian with this cell increased by 0.1, and the error of the Jacobian with this cell decreased by 0.1. Error was defined as 1 - Pearson correlation between the DMOS scores for all training images and the distance scores provided by the new Jacobian. If either modified Jacobian caused error to be reduced, the evaluated Jacobian with the smallest error was chosen as the new Jacobian. The algorithm iterated for 10,000 steps, which we subjectively determined to be the point at which error plateaued (a minimum was found) (**Supplemental Fig S 10**). This algorithm is visualized in **Fig 6**. Approach II succeeds despite the extreme simplicity of its optimization algorithm, which we find an argument for the power of the general approach. Regularization was avoided because it would be difficult to interpret the resultant Jacobian if its values are attributed to an unknown combination of correctness and regularization terms (e.g. sparsity).

In order to more easily interpret these results, we make the limiting assumption for approach II that a single Jacobian is applicable across the set of all images (or all examples of a scene category). Because this Jacobian is nonspecific to a particular type of image, we will be extracting only those partial differential equations applicable to all images.

For efficiency and because the Jacobian must be a symmetric matrix to guarantee that it will fulfill the tensor's desirable property of symmetry, the upper diagonal of the Jacobian is dependent only on the lower triangle. The diagonal is fixed to ones, accounting for the identity matrix in eqn. (10). Therefore, the error surface is $\frac{D\times(D-1)}{2}$ dimensional. Cells of the Jacobian were limited to the range [-1,1].

For the sake of computational complexity, we assume for approach II that $\mathbf{P}$ is local and uniform across the image (see **section 6** for a simple relief from this extension). This assumption is congruent with the low-level visual system, wherein relations between representations of topographical neighbors are one dominant component [48–52] [57] [58] [61]. It is also approximately true for the radial basis functions explored in approach I. Images were split into 8 x 8 pixel tiles. This enabled us to optimize a single 64 x 64 Jacobian (rather than a 65,536 x 65,536 Jacobian that has an untenable billion-dimensional error surface). The 64 x 64 Jacobian was used to compare each tile of an image, after which the tile distances were summed. This tiling approach is consistent with JPEG [1] [2], related compression methods [72], and other IQA measures (e.g. SSIM [12]). Sub-imaging greatly increased the number of data points used for training while simplifying the task.

Several considerations are relevant to this regression. We present results using a regression which allowed cells of the Jacobian to be negative. Negative cells in the Jacobian indicate that certain features contradict one another. Anecdotally, we found these results to be superior to non-negative regressions. Secondly, the solution of a gradient descent optimization may not be unique (the regression may find one of many local optima), so researchers would be prescient to run multiple regressions with different randomly initialized Jacobians. However, the high dimensionality of the error space means that it cannot be sufficiently sampled in a reasonable amount of time, so it is not naively practical to find a global minimum. We make anecdotal report that different random regressions (although all starting from the identity matrix) produce nearly identical results. This is likely because the implicit dimensionality of natural images is low [10] [73] [74] – most regions of image space are unpopulated. Thirdly, in cases not further covered here where most item-item perceived differences are known and other tricky conditions hold, one can extract an optimal Jacobian by reformulating the problem as a system of simple linear equations [72].

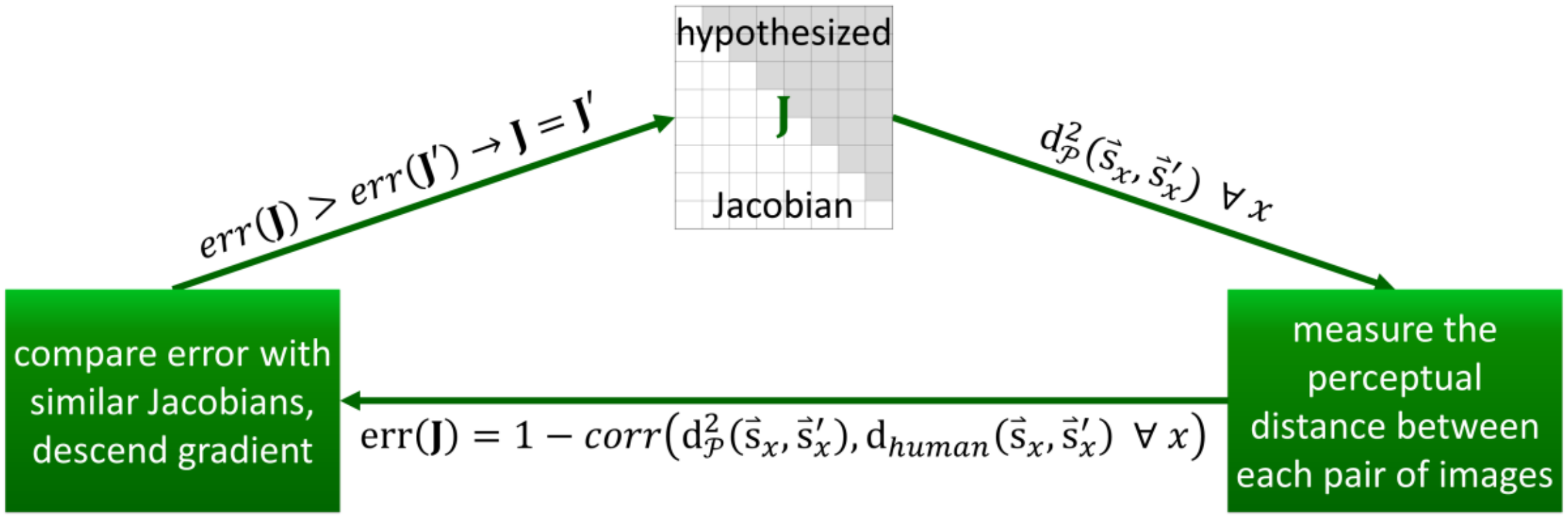

**Fig 6. Determination of an approach II Jacobian by way of regression**. A Jacobian is initialized. Second, the Jacobian is used to measure the distance between each image pair. Third, the error of this Jacobian is computed based on how well its distances correlate with human subject ratings. If this Jacobian produces reduced error, it is marked as the best working hypothesis. Finally, new Jacobians are generated with slight deviations from the best working hypothesis.

### 4.5 Analysis

When comparing correlation values in **Table 1**, we want to measure the statistical significance of pairwise differences between Spearman correlations with DMOS for competing models. One option is to perform X-fold crossvalidation of the original (non-degraded) images, and measure the significance across folds of the difference between two models. Due to approach II's regression runtime, we can at most acquire 5 folds. Five pairs is too few for the nonparametric Wilcoxon signed-rank test or Fisher-Pitman exact permutation test – the minimum possible p value is not significant. By contrast, the paired t-test (on Fisher z-transformed correlations) can yield any p value, but will be sensitive to outliers and variability in the results. Instead, we measured, for each of the two folds reported in **Table 1**, a 2-tailed Fisher r-to-z transformation, then took the mean across folds.

For within-dataset analyses, the training set of original images was randomly halved (preserving equal N among semantic categories), and two approach II Jacobians were independently optimized in a 2-fold crossvalidation. Each Jacobian was only used to predict images uninvolved in its training, and the two sets of test scores were pooled (without modification) for comparison with DMOS.

For across-dataset analyses, an approach II Jacobian was optimized on the entire training dataset and then used to predict the mean subject's DMOS score for each image of the testing dataset.

## 5 RESULTS AND PREDICTIVE CAPACITY

The question asked is: Do the tensors produced by approaches I and II explain most of the variance in human ratings not already explained by Euclidean measures? To test the ability of each connectivity pattern to predict human similarity judgments, we compare rank-order and linear correlation with human ratings. The outcomes of these analyses are apparent in the rank-order correlations of **Table 1**, which we emphasize in fairness to SSIM, whose relationship with DMOS is notably nonlinear. In linear correlation with the human ratings of SceneIQ Online, Pearson's $r = 0.4474$ for Euclidean; 0.6409 for approach I Gaussian $\sigma = 0.9$ pixels; 0.7559 for approach II (mean across 2 folds of data, all differ from chance $p << 0.001$). In **Supplemental Materials**, we also provide comparisons in terms of logistic regression.

Approach I with a Gaussian hypothesis improves drastically on the null Euclidean hypothesis, indicating that this connectivity plays a role (**Table 1**, **Fig 7**, adjusted $R^2$ of linear fit between Euclidean and DMOS on log-log axes increases from 0.1989 to 0.4217 when predictions from Gaussian $\sigma = 0.9$ are additionally included). More surprisingly, both approach I difference-of-Gaussians and approach II tensors consistently outperform the performance-driven SSIM.

It is important that the relationship between predictions and behavior be simple. Complicated relationships (e.g. logistic fits used in many IQA methods), or those with many parameters, require further explanation. **Fig 7** illustrates empirical DMOS scores from SceneIQ Online as predicted by Euclidean and approach II. In this case, the fit between approach II and empirical human ratings (DMOS) appears linear when plotted on log-log axes. In **Fig 7**, the four groupings of approach II ratings roughly correspond to the four JPEG quality levels in the dataset.

SceneIQ is evenly divided across eight semantic categories of visual scene. For approach II, a single Jacobian was identified for the set of all images. Nonetheless, this single transform constitutes the highest performing psychophysical predictor for most of the individual image categories (**Table 1**). It also can be seen to transfer well to the CSIQ dataset, where it still is among the best predictors, and still outperforms SSIM. Unexpectedly, perception is easier to predict with this approach *across* scene categories than within them.

**Table 1**
SPEARMAN CORRELATION WITH HUMANS

| | | Approach I | | | Approach II | | Euclidean | SSIM | p( Approach II = SSIM ) |
|---|---|---|---|---|---|---|---|---|---|
| | | Gauss σ = 0.9 | Gauss σ = 2 | Center Surround | SceneIQ All | SceneIQ Lab | | | |
| CSIQ (JPEG) | | 0.9541 | **0.9582** | 0.9578 | 0.9487 | 0.9484 | 0.8883 | 0.9224 | 0.0673 |
| CSIQ Revised | | 0.8496 | **0.8884** | 0.8868 | 0.8760 | 0.8715 | 0.5305 | 0.8104 | 0.0786 |
| TID2013 (JPEG) | | 0.6975 | 0.5391 | **0.9476** | 0.9311 | 0.9380 | 0.8782 | 0.9092 | 0.2630 |
| Toyama (JPEG) | | 0.4364 | 0.4554 | 0.5382 | 0.7605 | **0.7648** | 0.4240 | 0.6505 | 0.1594 |
| SceneIQ Online | All | 0.7223 | 0.6576 | 0.8295 | **0.8314** | - | 0.4738 | 0.7036 | ~0 * |
| SceneIQ Online | C | 0.6905 | 0.6042 | **0.8597** | 0.8574 | - | 0.5920 | 0.7550 | 0.000003 * |
| SceneIQ Online | F | 0.8004 | 0.7247 | 0.8356 | 0.8357 | - | 0.5238 | **0.8417** | 0.1464 |
| SceneIQ Online | H | 0.7096 | 0.6472 | **0.8616** | 0.8605 | - | 0.6355 | 0.7108 | ~0 * |
| SceneIQ Online | IC | 0.7952 | 0.7347 | 0.8382 | **0.8458** | - | 0.7207 | 0.8207 | 0.1902 |
| SceneIQ Online | M | 0.7396 | 0.6614 | 0.8380 | **0.8461** | - | 0.5585 | 0.7915 | 0.0113 |
| SceneIQ Online | OC | 0.6944 | 0.5828 | 0.8462 | **0.8470** | - | 0.5107 | 0.7761 | 0.0008 * |
| SceneIQ Online | S | 0.7954 | 0.7443 | 0.8446 | **0.8452** | - | 0.7531 | 0.8362 | 0.6585 |
| SceneIQ Online | TB | 0.6912 | 0.6015 | 0.8177 | **0.8182** | - | 0.6524 | 0.7751 | 0.0844 |
| SceneIQ Lab | | 0.7440 | 0.6570 | 0.8607 | - | **0.8655** | 0.5467 | 0.7567 | - |

Spearman correlation with humans (DMOS) for the presented models (columns) on several datasets (rows). Euclidean distance between images and SSIM are included for comparison. Except SceneIQ, correlations were calculated on entire datasets. SceneIQ Online and SceneIQ Lab correlations represent the mean across two random folds of original (non-degraded) images (see **section 4**). Approach II SceneIQ All was trained on each fold of SceneIQ Online and tested on the other fold (on other datasets, the reported correlation represents approach II fit to all SceneIQ Online, and tests the *generalization* of approach II trained on SceneIQ Online to a different dataset). For each dataset, the highest-performing model is marked in bold. All correlations were found to differ from zero with $p << 0.001$ via permutation test (after Bonferroni correction for 100 comparisons). The last column indicates the probability that SSIM and approach II SceneIQ All correlate equally with humans (***** indicates values below conservative Bonferroni threshold for 14 comparisons; some comparisons only in **Supplemental Materials**). P values were determined by 2-tailed Fisher r-to-z transformation of Spearman correlations (for SceneIQ Online and SceneIQ Lab, as measured for each fold then meaned). C = coast, F = forest, H = highway, IC = inside city, M = mountain, OC = open country, S = street, TB = tall building. “Lab” = laboratory validation.

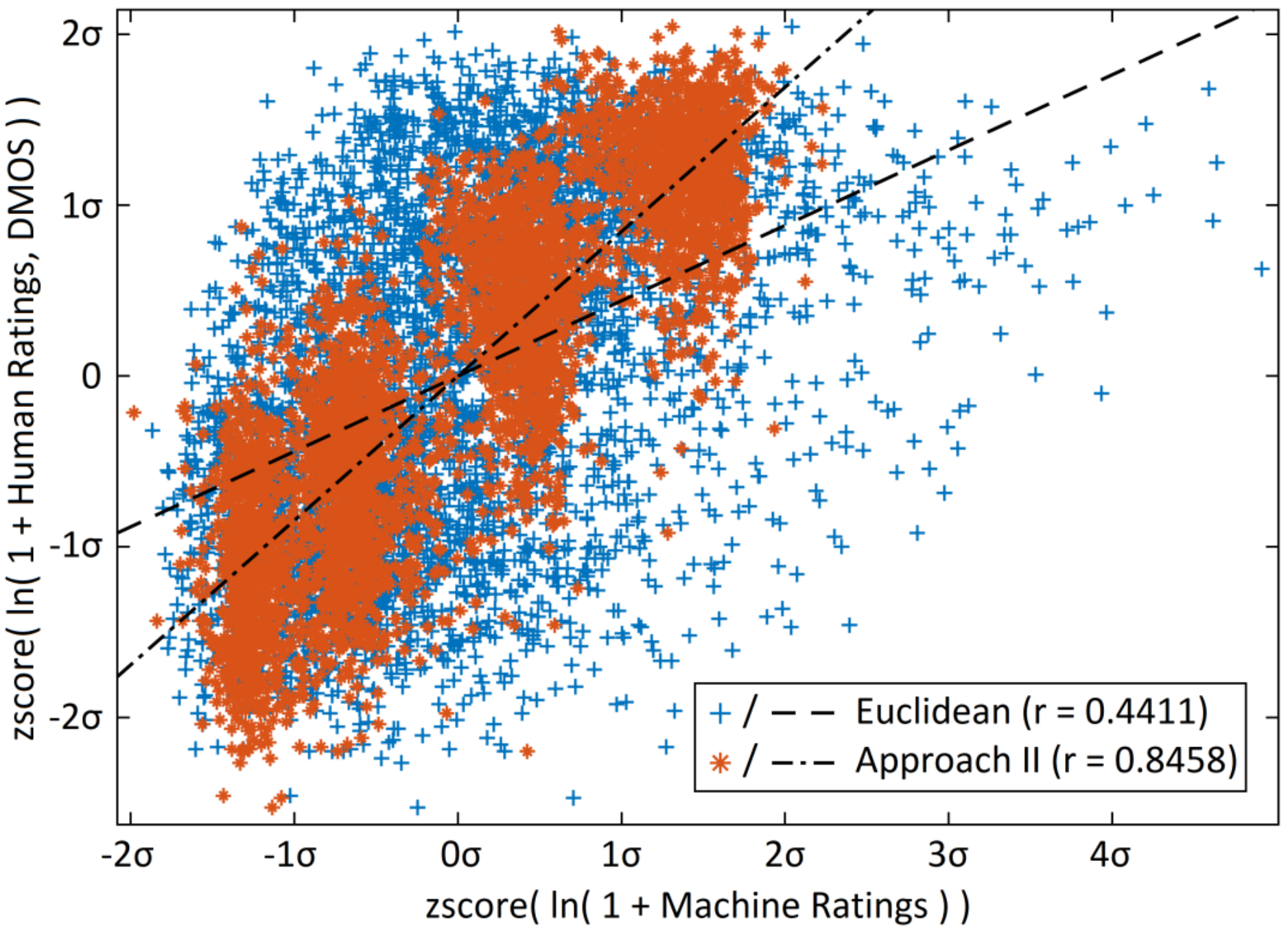

**Fig 7. Correlation of Euclidean and approach II with DMOS.** Half (first fold) of the full SceneIQ Online dataset on log-log axes. Machine ratings are on the X axis, while human DMOS ratings are on the Y. Euclidean and approach II ratings were z-scored separately so that they can be more usefully superimposed. We plot lines of best fit, and calculate Pearson's correlation on these log-log coordinates. (These linear correlations on log-log axes cannot be directly related to the correlations reported in text.) Linear axes, logistic fits, and comparisons with SSIM are available in **Supplemental Materials**.

## 6 Conclusions

Using well known data from neural connectivity in the early visual pathway, we showed the ability to predict simple human similarity judgments (IQA fidelity), suggesting that much of the variance in these psychophysical judgments may be explained by surprisingly simple neural principles. Moreover, the predictions routinely rivaled or outperformed those of standard approaches in the field.

Neural representations are non-Euclidean: relations among neighboring features distort image geometry. Approach I explicitly replicates several non-Euclidean connectivity patterns within the visual system. Despite incorporating only quite simple neural principles, analyses indicate that approach I alone can equal or outperform industry-standard IQA measures at predicting human behavior.

Approach II further added the (still simple) refinement of data-driven regression. It should be emphasized that the regression was accomplished with a very small body of empirical measures (a regression fit to just 80 images generalizes nearly as well as one fit to 2080; see **Supplemental Materials**), as opposed to typical big-data methods. The regression found a set of pixel-pixel relations that perceptually strained images such that comparison between them was humanlike; the results are in **Table 1** (see also **Supplemental Materials**).

The Jacobian matrices that result are directly interpretable in terms of connections among stimulus features. A given Jacobian directly represents hypotheses about connectivity in the early visual path, either from straightforward principled models (approach I), or derived from simply-regressed behavioral data (approach II). The framework is therefore flexible, from which many IQA approaches (such as SSIM) may be viewed as special cases when the restrictions of approaches I and II are lifted.

The more complex the relationship between predictions and empirical behavior, the more difficult it may be to unearth explanatory principles underlying predictive performance. It is hoped that the relatively straightforward methods forwarded here assist in the simplification of our understanding of similarity judgments.

We sought a formalism which quantified connectivity in units of input-input relations ($\partial \mathrm{u}_i / \partial \mathrm{s}_j$) (not input-output relations, $\mathrm{y}{=}f(\vec{\mathrm{s}})$, as is more typical in artificial neural network approaches). The findings suggest the potential of such formalisms to understand how patterns of individual associations yield the gestalt of an image percept, which is composed of many outputs working together, rather than in isolation. Such a formalism is aligned with many insights neuroscientists have acquired about connectivity. Further characterization of the types of candidate hypotheses is in progress.

To the extent that the newly introduced perceptual displacement field does strain stimuli toward their relative perceived locations, we directly predict that this will explain effects in other perceptual domains such as color constancy, visual filling-in, category-specific connectivity, change blindness, and visual illusions. We are actively investigating some of these domains in our lab. For instance, many disparate findings in visual crowding can be explained by a simple model that is almost identical to approach I, simply adding connectivity corresponding to increasing receptive field size that differs by degrees from fixation (Rodriguez, 2020, under review).

It should be emphasized that the IQA tasks, and corresponding datasets such as CSIQ and SceneIQ, are unlikely to capture higher-level aspects of perception. The approach presented here contains only simple connectivity profiles, speculated to be akin to early pathways, and cannot account for e.g. nonlinearities in categorical perception, top-down mechanisms in attention, or temporal dynamics in motion processing. Like the early visual stream [75–77] but unlike higher-level vision (such as face perception) [78] [79], our approaches are not invariant to translation and scaling (**Supplemental Materials**).

Those higher-level processes sit downstream from early vision. In the future, hierarchical and recurrent tensors may be important for capturing the above behavior. Approaches I and II may serve as valuable representations of the early visual pipeline, from which these more advanced models may draw. Human percepts are unlikely to derive from evenly weighted image regions, although weighting schemes have been previously proposed (e.g. [31]) and could eventually be applied here. Importantly, the brain may process distinct image regions differently depending on their content. Therefore, strain that is defined, not as constant, but as a function of the input, may become an even more important extension.

These extensions will pose new challenges. The computational complexity of fitting these models is substantial, and differential geometric approaches will eventually suffer from being underconstrained, so methods must be devised to reduce the degrees of freedom. However, many popular approaches to gradient descent, evolution, sampling, and back-propagation take the Euclidean assumption – that optimization parameters can be evaluated in isolation. In the non-Euclidean case, motion along one dimension of the error manifold changes the shape of the manifold in all directions. One interesting path of future investigation will be to revise optimization algorithms to account for the interrelations among features being modeled.

Our approaches are decidedly predictive IQA measures, which makes them useful as error measures in an algorithmic search for superior image compression formulae (e.g. [80] [81]). The presented predictive models may also reduce the need (in psychology, neuroscience, and software design) for high volume human data collection, which often prescribes strict standards for controls (e.g. [82] [83]).

The underlying framework presented here also may be considered for a broader set of objectives. In machine learning, artificial networks are seldom designed with predetermined connectivity, though biologically-informed connectivity in such networks has been advantageous [84]. A Jacobian can be cast as a predetermined weight matrix of a multi-layer perceptron (as in backpropagation, deep learning, and other forms). In the future, differential geometry may be used to simultaneously populate the weights of an artificial network based on a simple hypothesis. The approaches of this paper may assist in further engineering solutions to analyses of multivariate data containing spatially or functionally related features. Examples include the decoding of signals from brain electrode data, functional MRI, electrocorticography, computer vision, weather stations, or group buying behavior (collaborative filtering).

We pose the IQA problem as one of perceptually deforming image space, using Cartesian coordinates of pixels (bitmaps) as the axes – the units of association. It is entirely possible that this projection from image to percept is more difficult than from other input feature spaces / coordinate axes. Discrete cosine style transforms have performed well in other IQA measures (e.g. [85-87]), but in our case were found to be inferior (**Supplemental Materials**). One explanation is that there are less neighborhood relations among features in those spaces, making them sub-optimal for approaches based on feature-feature associations. Other coordinate axes (e.g. [62] [63] [88] [89]) may consist of features whose relations more simply and consistently predict behavior. Ideally, it may be possible to find an embedding of the images in which feature relations strain to explain perception in an optimal or parsimonious way.

If the proposed perceptual space is non-Euclidean, how do so many models built on Euclidean spaces and Cartesian coordinates flourish in applications? Under what criteria should Euclidean geometry be less preferred? One answer is that, when distances are sufficiently large, Euclidean distance is a close approximation of perceptual judgments. Thus perceptual evaluations deviate from Euclidean measures predominantly when distances are small, i.e., as perceptual distinctions become more challenging. This is another topic of ongoing study.

SceneIQ presents a new scale of dataset for IQA. It contains 69 times as many original (non-degraded) images as CSIQ, making analyses with higher-order statistics or high-degree-of-freedom regressions reliable. Images in SceneIQ are evenly distributed between eight semantic categories, enabling future semantic analyses. The original images are well characterized in terms of image statistics and perception (e.g. [68]), and these characterizations are available to future study in IQA. The viewing conditions and subject pool are naturalistically variable, yet have been validated in a more controlled laboratory setting. Perhaps, the ecological validity of real, variable viewing conditions makes these ratings an even more reliable benchmark than scores collected under highly controlled conditions. We hope the depth of this dataset makes it a valuable benchmark for the field.

## Acknowledgments

The authors thank James V Haxby, Jeremy Manning, and Josh Bongard for valuable discussions. This work was supported in part by grants N00014-15-1-2132 and N00014-16-1-2359 from ONR and N00014-15-1-2823 from DARPA.